\documentclass{article}
\pdfoutput=1

\usepackage{times}
\usepackage{graphicx}           % more modern
\usepackage{subfigure} 

\usepackage{natbib}
\usepackage{amsmath}            % basic features of AMS-LaTeX
\usepackage{amssymb}            % provide some more AMS-LaTeX symbols
\usepackage{array}              % extensions to tabular and array
\usepackage{algorithm}          % new float type for algorithms
\usepackage{algorithmic}        % bold keywords for pseudocode
\usepackage{booktabs}           % publication-quality tables

\usepackage{multicol}           % Multicolumn environment

\usepackage{tikz}               % New drawing package
\usetikzlibrary{bayesnet}       % Specialized environment for Bayesian Networks

\usepackage[accepted]{icml2014}

\usepackage{widetext}
\usepackage{arydshln}
\makeatletter

\renewcommand*\env@matrix[1][*\c@MaxMatrixCols c]{%
  \hskip -\arraycolsep
  \let\@ifnextchar\new@ifnextchar
  \array{#1}}
\makeatother

\newcommand{\NegBin}{\mathrm{NB}}  % Parametrize by mu and alpha
\newcommand{\Normal}{\mathcal{N}}  % Parametrize by mu and sigma^2
\newcommand{\Exponential}{\mathrm{Exponential}}
\newcommand{\E}{\mathbb{E}}
\newcommand{\Var}{\mathrm{Var}}
\newcommand{\Cov}{\mathrm{Cov}}
\newcommand{\Rset}{\mathbb{R}}     % set of reals

\newcommand{\bsymbol}[1]{\boldsymbol{#1}}

\newcommand{\bSigma}{\bsymbol{\Sigma}}

\newcommand{\bTheta}{\bsymbol{\Theta}}
\newcommand{\btheta}{\bsymbol{\theta}}

\icmltitlerunning{Effective Bayesian Modeling of Groups of Related
  Count~Time~Series}

\begin{document} 

\twocolumn[ \icmltitle{Effective Bayesian Modeling of Groups of
    Related Count~Time~Series}

% unless you've provided the [accepted] option to the icml2014
\icmlauthor{Nicolas Chapados}{chapados@apstat.com}
\icmladdress{ApSTAT Technologies Inc., 408-4200 Boul. St-Laurent, Montréal,
 QC, H2W 2R2, CANADA}
%\icmlauthor{Your CoAuthor's Name}{email@coauthordomain.edu}
%\icmladdress{Their Fantastic Institute,
%            27182 Exp St., Toronto, ON M6H 2T1 CANADA}

\icmlkeywords{Bayesian modeling, Time series forecasting}

\vskip 0.3in
]

\begin{abstract} 
Time series of counts arise in a variety of forecasting applications, for
which traditional models are generally inappropriate. This paper introduces
a hierarchical Bayesian formulation applicable to count time series that
can easily account for explanatory variables and share statistical strength
across groups of related time series. We derive an efficient approximate
inference technique, and illustrate its performance on a number of datasets
from supply chain planning.
\end{abstract} 

%%%%%%%%%%%%%%%%%%%%%%%%%%%%%%%%%%%%%%%%%%%%%%%%%%%%%%%%%%%%%%%%%%%%%%%%%%%%%%%

\section{Introduction}

Most classical time series forecasting models such as exponential smoothing
\cite{Hyndman:2008:Expsmooth} and ARIMA models \cite{Box:2008:time} assume
that observations are real-valued and can take on both positive and
negative values. In addition, the majority of classical approaches provide
normal predictive distributions, if they do so at all.  However, large
segments of the practice of forecasting---for instance in supply chain
planning---deal with time series that depart significantly from these
assumptions: series, for example, that may consist only of non-negative
integer observations, contain a large fraction of zeros, or are further
characterized by long runs of zeros interspersed by some large non-zero
values. In other words, the classical assumptions of conditional normality
are grossly violated.
Moreover, if multiple contemporaneous series are considered, common models
either treat them completely independently, or---as in vector
autoregressive models \cite{Box:2008:time}---attempt a more complex
multivariate modeling that captures short-range cross-correlations but
becomes unwieldy when managing hundreds of series; the common
scenario of ``weak coupling'' between related series (e.g. consumer
demand for a seasonal product at several stores of the same chain in a
given city, which could share seasonal behavior but not strong
cross-correlations) is not easily handled in classical modeling
frameworks.

\subsection{Motivating Applications}

The starting point for the present work lies in the intermittent demand
series that frequently occur in supply chain operations: these arise, for
example, in the demand for spare parts in aviation, or in the number of
``slow-moving'' items sold in retail stores \citep{Altay:2011:service}. In
addition to the non-negativity, integrality, skewness and high fraction of
zeros attributes already outlined, these time series are commonly quite
short: many weekly and monthly demand series encountered in practice may
consist of some 30 to 100 observations. This makes it crucial to allow some
information sharing across related series to more reliably capture posited
common effects such as seasonalities and the impact of causal determinants
(such as the market response to promotions or supply chain
disruptions). These are the modeling challenges that we address in this
paper.

\subsection{Related Work}

Most of the literature on intermittent demand forecasting relies on
relatively simple techniques, typically variants of
\citeauthor{Croston:1972}'s \citeyearpar{Croston:1972} method which
computes the expected demand in the next period as the ratio of the
expected non-zero demand to the expected non-zero-demand time interval,
both estimated by simple exponential smoothing.
%\footnote{\citet{Syntetos:2005:accuracy} proposed a correction to
%  the bias in Croston's approach, arising from the fact the expectation of
%  a ratio is not equal to the ratio of two expectations.} 
%
Croston's method produces point forecasts only;
\citet{Shenstone:2005:stochastic} studied variants with a proper stochastic
foundation that can produce predictive intervals, although with no
attempt to capture some known stylized facts of intermittent demand
patterns such as heavy tails \citep{Kwan:1991:demand-slow}.  It is only
with the recent work of \citet{Snyder:2012:forecasting} that a reasonably
modern formulation was proposed in terms of a state-space model and
distributional forecasts. This model %(in its better-performing variant)
still tracks the expected demand through an
exponential-smoothing update, but emits predictive distributions that
belong to the negative binomial family (described in the next
section); model parameters are estimated by maximum likelihood (ML). Despite the
evidence of improved accuracy against common benchmarks,
%presented by \citeauthor{Snyder:2012:forecasting}, 
this approach still exhibits a number of 
shortcomings: it is fundamentally univariate, does not easily allow 
explanatory variables, and the ML estimation framework does not reflect
model parameter uncertainty that arises with the very short series that are
common in practice.

%Despite admonitions from the forecasting community regarding the need to
%consider full distributional models to properly evaluate forecasting
%accuracy in a slow-moving context \citep{Hoover:2006}, 

\subsection{Contributions}

This paper makes the following three contributions: (\emph{i}) introducing
a hierarchical probabilistic state-space model that is a good match to
commonly-seen types of count data, allowing for explanatory variables and
permitting information sharing across groups of related series
(\S\ref{sec:model}); (\emph{ii}) introducing an effective inference
algorithm for computing posterior distributions over latent variables and
predictive distributions (\S\ref{sec:inference}); (\emph{iii}) assessing
the proposed approach's performance via a thorough experimental evaluation
(\S\ref{experimental-eval}).

%%%%%%%%%%%%%%%%%%%%%%%%%%%%%%%%%%%%%%%%%%%%%%%%%%%%%%%%%%%%%%%%%%%%%%%%%%%%%%%

\section{Hierarchical Model for Non-Negative Integer Time Series}
\label{sec:model}

We introduce the proposed model in several stages, starting with the basic
state-space structure and integer-valued observations
(\S\ref{subsec:core-model}), introducing explanatory variables and
structural zeros (\S\ref{subsec:explanatory-vars}), and finishing with the
hierarchical structure allowing information sharing across series
(\S\ref{subsec:hierarchy}).

\subsection{Core Model}
\label{subsec:core-model}

For a single non-negative time series, the model is expressed in
state-space form \citep[e.g.,][]{Durbin:2012:time}, where the latent state
$\eta_t$ at period $t=1,\ldots,T$ represents the log-expected value of the
non-negative integer observation $y_t$. Of those, we shall assume that the
first $T-h$ ($h\ge 1$) are observed, and the last $h$ constitute the future
values over which we would like to forecast. Representation in log-space
enforces the constraint that the process mean can never become
negative. The state space structure makes all observations independent of
each other conditionally on the latent state; here we assume that
observation $y_t$ is drawn from a negative binomial (NB) distribution with mean
$\exp\eta_t$ and size parameter $\alpha$ (which is independent of $t$),
\[
    y_t \sim \NegBin(\exp \eta_t, \alpha).
\]
The negative binomial distribution (parametrized by the mean $\mu$ instead
of the more usual probability of success in a trial;
e.g. \citealt{Hilbe:2011ua}) is given by
\begin{equation}
    P_{\NegBin}(y \mid \mu, \alpha) = 
        \binom{\alpha+y-1}{\alpha-1} \!\!
        \left(\frac{\alpha}{\mu+\alpha}\right)^\alpha \!\!
        \left(\frac{\mu}{\mu+\alpha}\right)^y\!\!,
\label{eq:NB-lkd}
\end{equation}
where $\binom{n}{m} \equiv \frac{\Gamma(n+1)}{\Gamma(m+1)\Gamma(n-m+1)}$ is
the binomial coefficient. The negative binomial is appropriate for count
data that is overdispersed with respect to a Poisson distribution (with the
variance greater than the mean); the size parameter $\alpha > 0$ governs
the level of overdispersion. The limiting case $\alpha\rightarrow\infty$
converges to a Poisson distribution.

\begin{figure}
\begin{tikzpicture}
  \node[obs]    (y1) {$y_1$};
  \node[obs   , right=0.6 of y1] (y2) {$y_2$};
  \node[obs   , right=0.6 of y2] (y3) {$y_3$};
  \node[obs   , right=0.6 of y3] (y4) {$y_4$};
  \node[latent, right=0.6 of y4] (y5) {$y_5$};
  \node[latent, right=0.6 of y5] (y6) {$y_6$};
  \node[latent, above=1.65 of y1] (eta1) {$\eta_1$};
  \node[latent, above=1.65 of y2] (eta2) {$\eta_2$};
  \node[latent, above=1.65 of y3] (eta3) {$\eta_3$};
  \node[latent, above=1.65 of y4] (eta4) {$\eta_4$};
  \node[latent, right=0.6 of eta4] (eta5) {$\eta_5$};
  \node[latent, right=0.6 of eta5] (eta6) {$\eta_6$};
  \node[latent, above=1.0 of y1, xshift=-0.5cm](alpha)   {$\alpha$};
  \node[latent, above=0.5 of eta1]              (tau0)  {$\tau_0$};
  \node[latent, above=0.5 of eta2]              (tau)   {$\tau$};
  \node[latent, above=0.5 of eta3, xshift=06mm] (mu)    {$\mu$};
  \node[latent, above=0.5 of eta5]              (phi)   {$\phi$};
  \edge{eta1}{y1};
  \edge{eta2}{y2};
  \edge{eta3}{y3};
  \edge{eta4}{y4};
  \edge{eta5}{y5};
  \edge{eta6}{y6};
  \edge{eta1}{eta2};
  \edge{eta2}{eta3};
  \edge{eta3}{eta4};
  \edge{eta4}{eta5};
  \edge{eta5}{eta6};
  \edge{tau0}{eta1};
  \edge{alpha}{y1,y2,y3,y4,y5,y6};
  \edge{tau,mu}{eta1,eta2,eta3,eta4,eta5,eta6};
  \edge{phi}{eta2,eta3,eta4,eta5,eta6};
\end{tikzpicture}
\vspace*{-2mm}
\caption{Basic state-space model for a single time series as a fully
  unfolded directed graphical model. Shaded nodes $\{y_1,\ldots,y_4\}$ are
  observed values; variables $y_5$ and $y_6$ are values to be
  forecasted. The dependence of the latent log-intensity process
  $\{\eta_t\}$ on all hyperparameters is made explicit.}
\vspace*{-3mm}
\label{fig:core-model}
\end{figure}
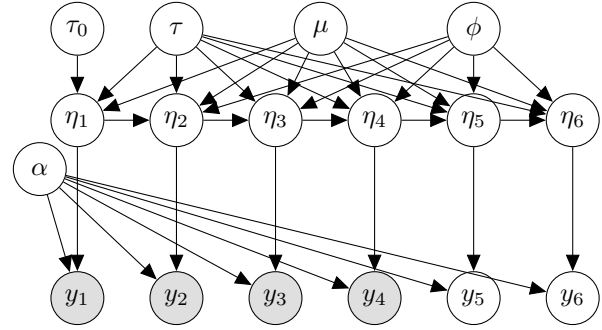

The dynamics of the process log-mean $\eta_t$ depend on the properties of
the time series being modeled. For a stationary series, a mean-reverting
autoregressive process is a sensible and tractable choice. In a supply
chain context, mean reversion intuitively means that the long-run expected
demand for an item, when projected far in the future, should fall back to a
constant level in spite of any past transient disturbances. We express
latent dynamics as an AR(1) process with normal innovations, with
\begin{equation}
\begin{aligned}
    \eta_1 &= \mu + \epsilon_1, \\[-0.5ex]
    \eta_t &= \mu + \phi(\eta_{t-1} - \mu) + \epsilon_t, && t > 1, \\[-0.5ex]
    \epsilon_1 &\sim \Normal(0, 1/\tau_0 + 1/\tau), \\[-0.5ex]
    \epsilon_t &\sim \Normal(0, 1/\tau),     && t > 1,
\end{aligned}
\label{eq:latent-process}
\end{equation}
where $\mu\in\Rset$ is the long-run level of mean reversion, $-1 < \phi <
1$ is the speed of mean reversion, $\tau>0$ is the precision of the process
innovations, and $\tau_0>0$ allows for additional variance in the initial
period. All $\epsilon_t$ are assumed mutually independent.
The model structure is depicted in graphical form in
Fig.~\ref{fig:core-model}. Forecasting in this model conceptually
proceeds in three steps: (\emph{i})~from the observed values of the time
series, we carry out inference over all unobserved model variables (the
clear nodes in Fig.~\ref{fig:core-model}); (\emph{ii})~using the inferred
process parameters ($\tau,\mu,\phi$), we project the latent dynamics into
the future to obtain a distribution over future values of the latent state
($\eta_5$ and $\eta_6$ in the figure); and (\emph{iii})~obtaining a
predictive distribution over future observations ($y_5$ and $y_6$ in the
figure). This description can easily be extended to accommodate
multivariate observations and latent states; the latent states would then
follow a vector autoregressive (VAR) process.

\subsection{Explanatory Variables and Structural Zeros}
\label{subsec:explanatory-vars}

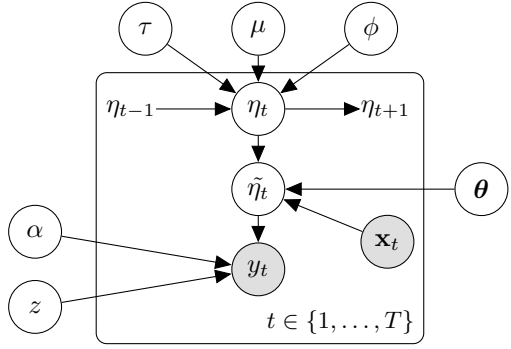
\begin{figure}
\centering
\begin{tikzpicture}
  \node[obs]                         (yt)     {$y_t$};
  \node[latent, above=0.35 of yt]     (tetat)  {$\tilde{\eta_t}$};
  \node[latent, above=0.35 of tetat]  (etat)   {$\eta_t$};
  \node[const , left=of etat]        (etatm1) {$\eta_{t-1}$};
  \node[const , right=of etat]       (etatp1) {$\eta_{t+1}$};
  \node[latent, left=2.25 of yt, yshift=+5mm]       (alpha){$\alpha$};
  \node[latent, left=2.25 of yt, yshift=-5mm]       (z)    {$z$};
  \node[latent, above=0.35 of etat, xshift=-15mm]   (tau)  {$\tau$};
  \node[latent, above=0.35 of etat]                 (mu)   {$\mu$};
  \node[latent, above=0.35 of etat, xshift=+15mm]   (phi)  {$\phi$};
  \node[latent, right=2.25 of tetat]               (theta){$\btheta$};
  \node[obs, right=of tetat, yshift=-7mm]          (xt)   {$\mathbf{x}_{t}$};

  % Edges
  \edge{etat}{etatp1,tetat};
  \edge{tetat}{yt};
  \edge{etatm1,tau,mu,phi}{etat};
  \edge{alpha,z}{yt};
  \edge{xt,theta}{tetat};

  % Plates
  \plate {} {(yt)(xt)(etat)(etatm1)(etatp1)} { $t \in \{1,\ldots,T\}$ };
\end{tikzpicture}
\vspace*{-3mm}
\caption{Incorporating explanatory variables $\mathbf{x}_t$ and
  zero-inflation $z$ into the model, where the plate indicates repetition
  over time periods $t$. For brevity, license is taken to omit depiction of
  the initial and final latent log-intensity, as well as representing the
  unshaded $\{y_t\}$ over the forecasting period.}
\vspace*{-3mm}
\label{fig:explanatory-vars}
\end{figure}

Explanatory variables (which can include seasonal terms, as well as factors
that causally impact the observed time series) can be incorporated by
viewing them as \emph{local forcing terms} that temporarily shift the
location of the latent process mean. This is illustrated in
Fig.~\ref{fig:explanatory-vars}. We assume that explanatory variables at
period $t$, $\mathbf{x}_t \in \Rset^N$, are always observed, non-stochastic
and known ahead of time (so that we know the future values of
$\{\mathbf{x}_t\}$ over the forecasting horizon). They are linearly
combined through regression coefficients $\btheta$ to additively shift the
latent $\eta_t$, yielding an effective log-mean $\tilde{\eta}_t$,
\begin{equation}
  \tilde{\eta}_t = \eta_t + \mathbf{x}_t'\btheta,
  \label{eq:eta-tilde}
\end{equation}
where the $\{\eta_t\}$ follows the same AR(1) process as previously. In
latent (log) space, the addition operation corresponds to a multiplicative
impact of explanatory variables on the process mean in observation space,
which is often a good fit to the underlying data generating process
(e.g. seasonalities, or consumer response to promotions or special
events). It also makes the regression coefficients $\btheta$ relatively
independent of the scale of the series, and makes it easier to share
information across multiple time series as described in
\S\ref{subsec:hierarchy}.

In many real-world series, one observes an excess of zero values compared
to the probability under the NB \citep[e.g.,][]{Lambert:1992vk}: this can
arise for structural reasons in the underlying process (e.g., out-of-stock
items or supply chain disruptions in a retail store context, both of which
would override the natural consumer demand modeled by the NB). For these
reasons, we add extra unconditional mass at zero, yielding so-called
\emph{zero-inflated} NB observations,
\begin{equation}
  y_t \sim z\,\delta_0 + (1-z)\,\NegBin(\exp\tilde{\eta}_t, \alpha),
  \label{eq:obs-model}
\end{equation}
where $\delta_0$ represents unit probability mass at zero and $z \in [0,1]$
is the probability of structural zero. We assume a $\mathrm{Beta}(\tfrac{1}{2},
\tfrac{1}{2})$ prior for $z$. This is our final observation model.

\subsection{Sharing Information Across Multiple Time Series}
\label{subsec:hierarchy}

Finally, we allow for a group of $L$ related time series to share
information, particularly in the form of a shared hyperprior over
regression coefficients and latent process characteristics. In the spirit
of hierarchical models studied in statistics \citep{Gelman:2007:data} and
machine learning \citep{TehJordan:2010:BN,Fox:2010:sharing}, we let those
parameters (for all time series $\ell \in \{1,\ldots,L\}$ that belong to
the group being modeled simultaneously) share common parents, as
illustrated on Fig.~\ref{fig:hierarchical-model}. A new plate iterates over
the series-level parameters, which inherit as follows from ``global''
parameters shared across all time series:
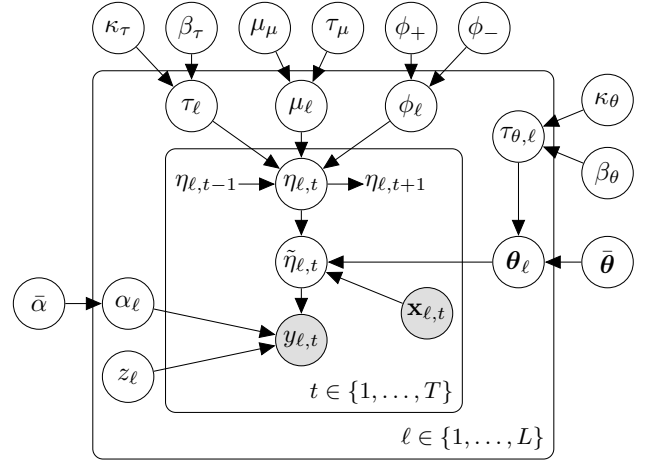
\begin{figure}[t!]
\resizebox{\columnwidth}{!}{%
\begin{tikzpicture}
  \node[obs]                                    (ylt)       {$y_{\ell,t}$};
  \node[latent, above=0.35 of ylt]               (tetalt)    {$\tilde{\eta}_{\ell,t}$};
  \node[latent, above=0.35 of tetalt]            (etalt)     {$\eta_{\ell,t}$};
  \node[const,  left=0.5 of etalt]              (etaltm1)   {$\eta_{\ell,t-1}$};
  \node[const,  right=0.5 of etalt]             (etaltp1)   {$\eta_{\ell,t+1}$};
  \node[latent, left=1.65 of ylt, yshift=+5mm]  (alphal)    {$\alpha_\ell$};
  \node[latent, left=1.65 of ylt, yshift=-5mm]  (zl)        {$z_\ell$};
  \node[latent, above=0.35 of etalt,xshift=-15mm](taul)      {$\tau_\ell$};
  \node[latent, above=0.35 of etalt]             (mul)       {$\mu_\ell$};
  \node[latent, above=0.35 of etalt,xshift=+15mm](phil)      {$\phi_\ell$};
  \node[latent, left=0.5 of alphal]             (alphab)    {$\bar{\alpha}$};
  \node[latent, above=0.35 of taul, xshift=-10mm](kappat)    {$\kappa_\tau$};
  \node[latent, above=0.35 of taul]              (betat)     {$\beta_\tau$};
  \node[latent, above=0.35 of mul, xshift=-5mm]  (mumu)      {$\mu_{\mu}$};
  \node[latent, above=0.35 of mul, xshift=+5mm]  (taumu)     {$\tau_{\mu}$};
  \node[latent, above=0.35 of phil]              (phipl)     {$\phi_{+}$};
  \node[latent, above=0.35 of phil, xshift=+10mm](phimi)     {$\phi_{-}$};
  \node[latent, right=2.25 of tetalt]           (thetal)    {$\btheta_\ell$};
  \node[latent, above=of thetal]                (tault)     {$\tau_{\theta,\ell}$};
  \node[latent, right=0.5 of thetal]            (thetab)    {$\bar{\btheta}$};
  \node[latent, right=0.5 of tault, yshift=+5mm](kappath)   {$\kappa_\theta$};
  \node[latent, right=0.5 of tault, yshift=-5mm](betath)    {$\beta_\theta$};
  \node[obs, right=of tetalt, yshift=-7mm]      (xlt)       {$\mathbf{x}_{\ell,t}$};

  % Edges
  \edge{etalt}{etaltp1,tetalt};
  \edge{etaltm1,taul,mul,phil}{etalt};
  \edge{tetalt,alphal,zl}{ylt};
  \edge{alphab}{alphal};
  \edge{kappat,betat}{taul};
  \edge{mumu,taumu}{mul};
  \edge{phipl,phimi}{phil};
  \edge{thetal,xlt}{tetalt};
  \edge{thetab,tault}{thetal};
  \edge{kappath,betath}{tault};

  % Plates
  \plate {time} {
    (ylt)(xlt)(tetalt)(etalt)(etaltm1)(etaltp1)
  } { $t \in \{1,\ldots,T\}$ };
  \plate {loc} {
    (alphal)(taul)(mul)(phil)
    (thetal)(tault)
    (time.south east)
  } { $\ell \in \{1, \ldots, L \}$ };
\end{tikzpicture}}
\vspace*{-5mm}
\caption{Model with hierarchical structure, where the outer plate indicates
  repetition across several time series $\ell$. Information sharing across
  series is achieved by ``global'' hyperparameters located outside all
  plates, in particular the regression coefficient prior $\bar{\btheta}$.}
\vspace*{-3mm}
\label{fig:hierarchical-model}
\end{figure}
\begin{align*}
    \alpha_{\ell}   &\sim \Exponential(\bar{\alpha}),                       &
    \mu_{\ell}      &\sim \Normal(\mu_\mu, 1/\tau_\mu),                     \\[-0.5ex]
    \tau_{\ell}     &\sim \Gamma(\kappa_\tau, \beta_\tau),                  &
    \tau_{0,\ell}   &\sim \Gamma(\kappa_{0,\tau}, \beta_{0,\tau}),          \\[-0.5ex]
    \phi_{\ell}     &\sim \mathrm{Beta}(\phi_{+} + \phi_{-}, \phi_{-}),     &
    \btheta_{\ell}  &\sim \Normal(\bar{\btheta}, \tfrac{1}{\tau_{\theta,\ell}}I),  \\[-0.5ex]
    \tau_{\theta,\ell} &\sim \Gamma(\kappa_\theta, \beta_\theta),
\end{align*}
where $\Gamma(a,b)$ represents the gamma distribution with shape parameter
$a$ and scale parameter $b$, and $\mathrm{Beta}(a,b)$ is the beta
distribution with shape parameters $a$ and $b$. The series-level parameters
(variables in plate $\ell$ on Fig.~\ref{fig:hierarchical-model}) all have
the same meaning as previously, except that the $\ell$ index makes them
dependent on a specific time series. The hyperpriors that are used for the
global parameters are given in the appendix. The latent
dynamics of $\{ \eta_{\ell,t} \}$ and the observation model are the same as
in the previous sections. For convenience, we shall denote all ``global''
variables in Fig.~\ref{fig:hierarchical-model} (except $\mu_{\mu}$, for
reasons to be made clear shortly) by $\bTheta_G=\{ \bar{\alpha},
\tau_{\mu}, \kappa_\tau, \beta_\tau, \kappa_{0,\tau}, \beta_{0,\tau},
\kappa_\theta, \beta_\theta, \phi_{+}, \phi_{-}, \bar{\btheta} \}$, all
series-$\ell$-level variables (except $\mu_\ell$) by $\bTheta_\ell=\{
z_l, \alpha_{\ell}, \tau_{\ell}, \tau_{0, \ell}, \phi_{\ell}, \btheta_{\ell},
\tau_{\theta,\ell} \}$, and all latents over which we should do inference
by $\bTheta = \bTheta_G \cup \{ \bTheta_\ell \} \cup \{ \mu_{\mu},
\mu_\ell, \eta_{\ell,t} \}$.

In the remainder of this paper, we call this model the \emph{hierarchical
  negative-binomial state space} (\textbf{H-NBSS}) model. It must be
stressed that this model assumes that all time series in the group are
conditionally independent given the series-level parameters. In particular,
the model does not allow expressing observation-level cross-correlations
across different time series $\ell_i \ne \ell_j$, except through common
effects coming from explanatory variables.\footnote{And except, of course,
  if the observations $y_{\ell,t}$ are themselves multivariate, which is
  outside the scope of this paper.}

%%%%%%%%%%%%%%%%%%%%%%%%%%%%%%%%%%%%%%%%%%%%%%%%%%%%%%%%%%%%%%%%%%%%%%%%%%%%%%%

\section{Inference}
\label{sec:inference}

Due to the non-conjugacy between the zero-inflated negative-binomial
likelihood and the normal latent log-mean process prior (and the general
difficulty of finding useful conjugate priors for negative-binomial
likelihoods), inference in the H-NBSS model does not have an analytically
tractable solution. One must resort to approximation techniques, which
fall, broadly speaking, into two families: deterministic and stochastic
methods \citep{Barber:2011:bayesian}. Of the deterministic approaches,
early examples include assumed density filtering
\citep[ADF,][]{Maybeck:1979vh}, which is a sequential projection approach,
as well as numerical integration schemes such as the piecewise
approximation of \citet{Kitagawa:1987:state-space}. More recently, the
expectation propagation (EP) algorithm of \citet{Minka:2001:PhD}, a
generalization of ADF, has proved successful in a number of non-linear
filtering and smoothing problems
\citep{Heskes:2002:expectation,Yu:2006:expectation,Deisenroth:2012:EP}.  As
to stochastic approaches, they can take the form of variants of Gibbs
sampling, such as the recursive forward-filtering backward-sampling (FFBS)
algorithm \cite{Robert:1999:convergence,Scott:2002:bayesian} as well as
sequential Monte Carlo techniques such as particle filtering
\citep[reviewed by][]{Doucet:2001:sequential}. \citet{Durbin:2000uq}
present an alternative approach based on importance sampling and antithetic
variables.  Static hierarchical regression models have been widely studied
in the statistics literature \citep{Gelman:2007:data}, where typical
inference techniques rely on block Gibbs sampling. Dynamic hierarchical
models have been less commonly studied, with the notable exception of the
nonparametric Bayesian model of \citet{Fox:2010:sharing}, who use an
efficient form of the Metropolis-Hastings algorithm for inference

It is imperative to contrast the benefits of a proposed algorithm to the
requirements of forecasting practice: for instance, in a supply chain
context, it is routine business to process tens to hundreds of millions of
time series on a daily or weekly basis.\footnote{For example, a large
  department retail store may sell 100K different items (Stock Keeping
  Units, SKUs); a chain with 1000 stores would then require periodic
  forecasts for 100M series.} Despite their inadequacies, practitioners
still rely on very computationally simple methods such as exponential
smoothing for the vast majority of their tasks. Needless to say, for a
forecasting approach to have an impact in practice, its accuracy benefits
must justify its computational cost. This seems to rule out all stochastic
algorithms, as well as many deterministic ones such as EP.

We shall argue that for the H-NBSS model, a Gaussian approximation of the
latent variables at their posterior mode, known as the Laplace
approximation \citep{Bishop:2006:pattern}, yields near-optimal performance
at extremely attractive computational cost compared to the
alternatives. One reason to expect good performance is that most of the
important (for forecasting) latent variables in the model ($\{ \mu_{\mu},
\mu_\ell, \eta_{\ell,t} \}$) have a conditionally normal prior; their
posterior is nearly always close to normality despite the non-linear
likelihood.

% -- despite the non-linear likelihood function, a big portion of the (prior)
% latent structure of the model is made up of conditionally-linear Gaussian
% variables. these constitute a Gaussian Markov Random Field (GMRF)
% . 

\subsection{Posterior Calculation}

The Laplace approximation requires to calculate the log-posterior
probability up to an additive constant,
\begin{equation}
  \log P(\bTheta \mid \mathbf{Y}) = 
    \log P(\bTheta) + 
        \log P(\mathbf{Y} \mid \bTheta) + C,
  \label{eq:posterior}
\end{equation}
where $\mathbf{Y} = \{y_{\ell,t}\}_{t=1}^{T-h}$ is the set of all observed
series values in all groups and $C$ is an unknown (and for the Laplace
approximation, unimportant) constant. The log-likelihood term $\log
P(\mathbf{Y} \:|\: \bTheta)$ is derived straightforwardly from the
observation model~\eqref{eq:obs-model} along with the negative binomial
probability distribution~\eqref{eq:NB-lkd}. The first term---the
log-prior---decomposes into global-, series- and observation-level terms,
\vspace*{-3mm}
\begin{multline*}
    \log P(\bTheta) = \log P(\bTheta_G) + 
                      \sum_{\ell=1}^L \log P(\bTheta_\ell \mid \bTheta_G) + \\[-2ex]
      \log P(\mu_\mu) + 
      \sum_{\ell=1}^L \log P\big(\mu_\ell, \{\eta_{\ell,t}\} \mid 
                                 \bTheta_\ell, \mu_\mu\big).
\end{multline*}
The second line of this equation expresses a prior over jointly
normally-distributed variables with a highly structured (and very sparse)
precision matrix, a Gaussian Markov random field
\citep[GMRF, see][]{Rue:2005:GMRF}, to which we now turn.

\subsection{GMRF Prior}

For the single time series process described in \eqref{eq:latent-process},
assuming that the initial $\eta_1$ has the long-run process distribution
with precision of $1/(\tau(1-\phi^2))$,\footnote{So that $\tau_0$ would
  equal $\phi^2/(\tau(1-\phi^2))$.} the joint prior over the latent process
$\{\eta_t\}$ along with normally-distributed long-run mean $\mu$ (having
prior precision $\tau_\mu$) is normally distributed with a
\emph{tridiagonal precision matrix} $Q$, except for the last row and column
(corresponding to $\mu$), that follows the pattern
\begin{equation*}
Q=
\scriptsize
\renewcommand{\arraystretch}{0.75}%
\begin{bmatrix}
 \tau  & -\tau  \phi  & 0 & 0 & 0 & \tau  \tilde{\phi} \\
 -\tau  \phi  & \tau(\phi^2+1) & \ddots  & 0 & 0 & -\tau\tilde{\phi}^2 \\
 0 & -\tau  \phi  & \ddots & -\tau  \phi  & 0 & \vdots  \\
 0 & 0 & \ddots  & \tau(\phi^2+1) & -\tau  \phi  & -\tau\tilde{\phi}^2 \\
 0 & 0 & 0 & -\tau  \phi  & \tau  & \tau  \tilde{\phi} \\
 \tau\tilde{\phi} & -\tau\tilde{\phi}^2 & \cdots  & -\tau\tilde{\phi}^2 & \tau\tilde{\phi} & \tau_\mu + \tau \psi_T \\
\end{bmatrix},
\end{equation*}
where $\tilde{\phi} \equiv \phi-1$ and $\psi_T \equiv T - 2(T-1)\phi +
(T-2)\phi^2$, where $T$ is the number of observations. The determinant of
this matrix is $\tau^T \tau_\mu(1-\phi^2)$.\footnote{These results were
  obtained by direct symbolic matrix inversion in Mathematica, and can be
  verified to yield the identity matrix when multiplying $Q$ with
  the joint $\{\eta_t,\mu\}$ covariance matrix.} The sparsity of $Q$ makes it
extremely fast to compute the process prior term. 

When considering the hierarchical model of section~\ref{subsec:hierarchy},
a similar sparsity pattern holds: the joint precision matrix across all
variables that belong to the GMRF prior has block-diagonal structure, with
one sub-matrix like $Q$ for each time series, and a final row/column
linking the series means $\{\mu_\ell\}$ to the global mean
$\mu_\mu$. Details are given in the appendix.

\subsection{Optimization and Predictive Distribution}

Maximization of~\eqref{eq:posterior} over $\bTheta$ can be carried out
efficiently using Quasi-Newton methods such as L-BFGS
\citep{Nocedal:1999:optim-book}.\footnote{Local optima are not a problem in
  practice as long as the $\{ \eta_{\ell,t} \}$ are suitably initialized; a
  reasonable initialization for $\eta_{\ell,t}, 1 \le t \le T-h$ can be
  taken as the midpoint between between $\log y_{\ell,t}$ and the mean of
  log-values for series $\ell$, $m_\ell$. Initial values for $T-h+1 \le t
  \le T$ can be taken to be $m_\ell$.} Let $\hat\bTheta$ be the maximizing
value and $\hat\bSigma$ the inverse of $H_{\hat\bTheta}$, the Hessian
matrix of~\eqref{eq:posterior} evaluated at $\hat\bTheta$. The Laplace
approximation posits that the posterior distribution over $\bTheta$ is
jointly normal with mean $\hat\bTheta$ and covariance matrix
$\hat\bSigma$. Due to the structure of the GMRF prior, matrix
$H_{\hat\bTheta}$ is nearly block-diagonal except for the variables that
belong to $\bTheta_G$. This makes it efficient to compute $\hat\bSigma$ by
sparse matrix solvers \citep[e.g.,][]{Davis:2006:direct}.

From the graphical model structure (Fig.~\ref{fig:hierarchical-model}) and
the observation model~\eqref{eq:obs-model}, the predictive distribution
over a future value $y_{\ell,t}, t \ge T-h+1$ depends only on the
distributions of $\tilde{\eta}_{\ell,t}, \alpha_\ell$ and $z_\ell$. In
practice, the posterior uncertainty over $z_\ell$ and $\alpha_\ell$ is
small and can be neglected. From the observation model, the posterior
distribution over $y_{\ell,t}$ can be obtained by integrating out
$\tilde{\eta}_{\ell,t}$,
\begin{align*}
  P(y_{\ell,t} \mid \mathbf{Y}) &=
  \int_{-\infty}^\infty \! P(y_{\ell,t} \mid \exp\tilde{\eta}_{\ell,t}) \,
       P(\tilde{\eta}_{\ell,t} \mid \mathbf{Y}) \,
       \mathrm{d}\tilde{\eta}_{\ell,t}\\
  &= P\Bigl(y_{\ell,t} \:\Bigl|\:
        \int_{-\infty}^\infty \! \exp\tilde{\eta}_{\ell,t}
                                 P(\tilde{\eta}_{\ell,t} \mid \mathbf{Y}) \,
                                 \mathrm{d}\tilde{\eta}_{\ell,t}
      \Bigr.\Bigr),
\end{align*}
where a key use of the well-known summation property of the negative
binomial is made, wherein for IID variables $X_i \sim \NegBin(\mu_i,
\alpha)$, we have $\sum_i X_i \sim \NegBin\Bigl(\sum_i \mu_i,
\alpha\Bigr)$, and we assume that the summation converges to an integral in
the limit. Hence, only the posterior expectation of
$\exp\tilde{\eta}_{\ell,t}$ is needed, which is readily obtained as
\[
  \E\big[ \exp\tilde{\eta}_{\ell,t} \mid \mathbf{Y} \big] = 
    \exp\Bigl(
       \E\left[ \tilde{\eta}_{\ell,t} \mid \mathbf{Y} \right] + 
       \tfrac{1}{2}\Var\left[ \tilde{\eta}_{\ell,t} \mid \mathbf{Y} \right]
    \Bigr),
\]
since $\tilde{\eta}_{\ell,t}$ has a normal posterior under the Laplace
approximation. From~\eqref{eq:eta-tilde},
\[
  \E\left[ \tilde{\eta}_{\ell,t} \mid \mathbf{Y} \right] =
      \E[\eta_{\ell,t} \mid \mathbf{Y}] + 
      \mathbf{x}_{\ell,t}'\E[\btheta_\ell \mid \mathbf{Y}],
\]
where the conditional posteriors for $\eta_{\ell,t}$ and $\btheta_\ell$ are
directly available in $\hat\bTheta$. Similarly, the posterior variance for
$\tilde{\eta}_{\ell,t}$ is
\begin{multline*}
  \Var\left[ \tilde{\eta}_{\ell,t} \mid \mathbf{Y} \right] =
      \Var[\eta_{\ell,t} \mid \mathbf{Y}] + \\
      \mathbf{x}_{\ell,t}'\Var[\btheta_\ell \mid \mathbf{Y}]\mathbf{x}_{\ell,t} + 
      2 \mathbf{x}_{\ell,t}'\Cov[\eta_{\ell,t}, \btheta_\ell \mid \mathbf{Y}]
\end{multline*}
where the variances and covariances on the right-hand side are from
$\hat\bSigma$.

\subsection{Accuracy Compared to MCMC}

\begin{figure}
\resizebox{\columnwidth}{!}{\includegraphics{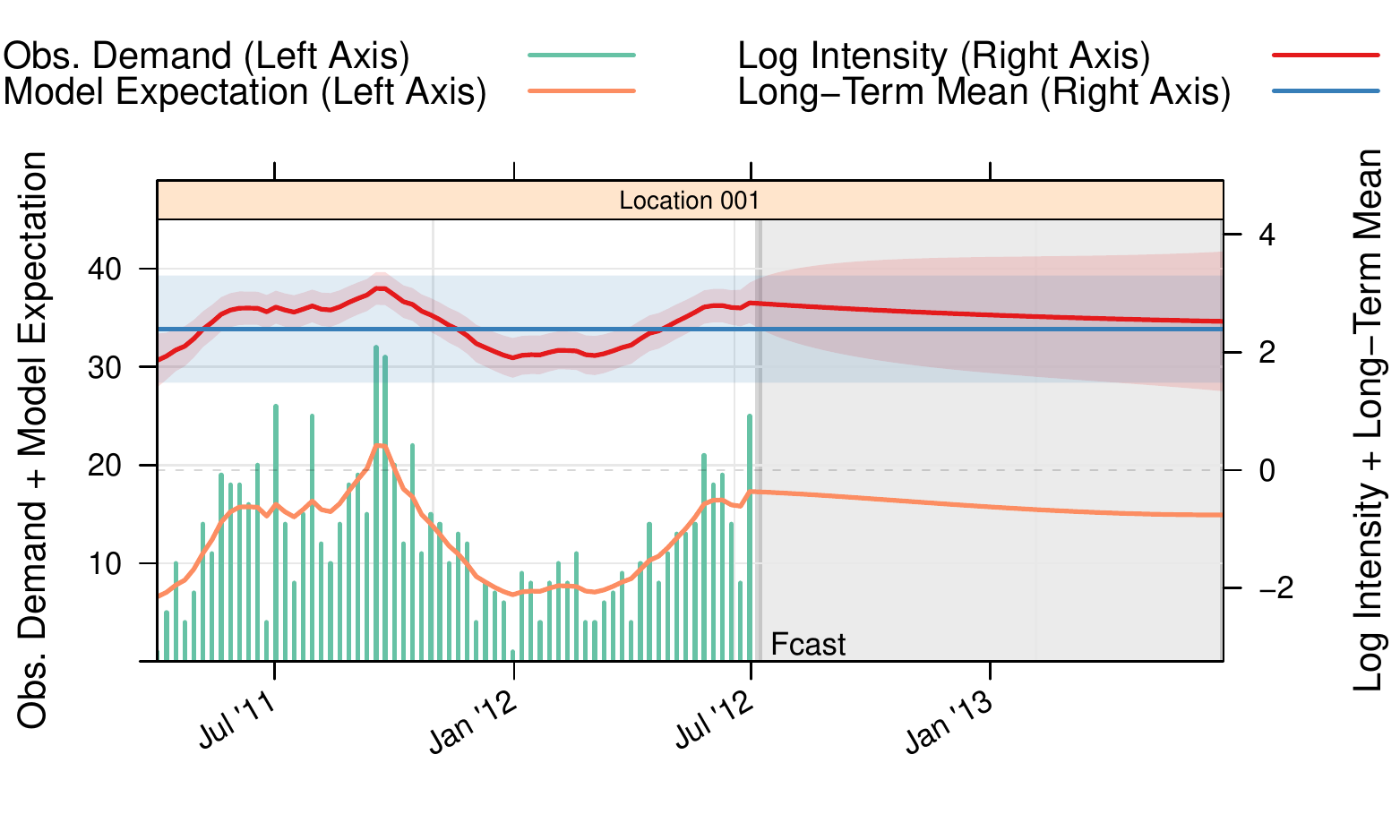}}%
\vspace*{-5mm}
\resizebox{\columnwidth}{!}{\includegraphics{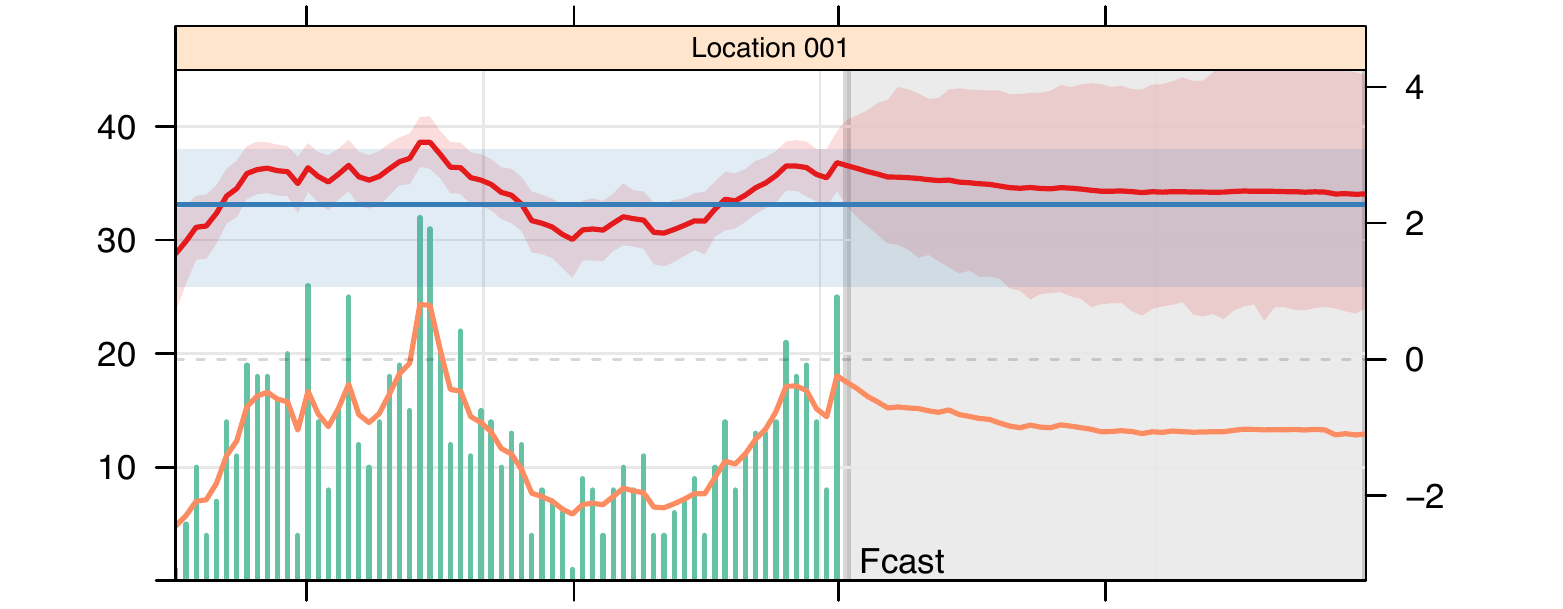}}
\vspace*{-6mm}
\caption{Comparison of inference results between the \textbf{Laplace
    approximation (top)} and \textbf{MCMC (bottom)}. The observed demand
  (green vertical lines) and model expectation of $y_t$ (continuous orange
  line) should be interpreted according to the \textbf{left axis}. The
  latent variables $\eta_t$ (red curve, denoted ``log-intensity'') and
  level of mean-reversison (blue horizontal line) should be interpreted
  according to the \textbf{right axis}. Shaded areas are 95\% credible
  intervals. The posteriors between the two approaches are close.}
\vspace*{-5mm}
\label{fig:Laplace-vs-MCMC}
\end{figure}

Ultimately, the validity of approximate inference is predicated on its
empirical performance, which is evaluated in the next section. Here, we
graphically contrast on Fig.~\ref{fig:Laplace-vs-MCMC} the inference
results for a single series between the Laplace approximation outlined
previously and an equivalent model computed with Markov chain Monte Carlo
(MCMC). The latter is implemented in the Stan modeling language
\citep{Stan-software:2013}, which uses the ``no-U-turn'' variant
\citep{Hoffman-Gelman:2013} of Hamiltonian Monte Carlo (HMC). We combined
the results of four independent chains, each run with 1500 burn-in
iterations followed by 18500 sampling iterations. Overall, we note the
similarity of the posterior distributions between the two approaches,
although the Laplace approximation slightly underestimates the posterior
variance in $\eta_t$ over the forecast horizon (the region denoted
``Fcast'' in the plots) compared with MCMC. Should additional accuracy be
required in the Laplace approximation, one could turn to a numerical
integration technique for hyperparameters of models equipped with GMRF
priors, the so-called integrated nested Laplace approximation
\citep[INLA,][]{Rue:2009:approximate}.
%and were studied recently in the machine learning community \cite{Han:2013:non-factorized}.

Of significant importance for practical applications, however, is
computational time. For the results illustrated in
Fig.~\ref{fig:Laplace-vs-MCMC}, whereas our implementation of the Laplace
approximation (coded in the interpreted language R) converges in a few
seconds, a roughly equivalent run of the MCMC sampler takes 30 times longer
(and Stan is a very efficient engine, compiling the model into C++ code
with analytical gradient computation for HMC). As will be clear from the
experimental results, the additional computational cost of MCMC does not
translate into a performance advantage in forecasting.

%%%%%%%%%%%%%%%%%%%%%%%%%%%%%%%%%%%%%%%%%%%%%%%%%%%%%%%%%%%%%%%%%%%%%%%%%%%%%%%

\section{Experimental Evaluation}
\label{experimental-eval}

%\subsection{Methodology and Data}

We evaluate model performance on three datasets obtained from supply chain
operations. The first one (RAID) is the sales of bug spray at 26 locations
of a major US retailer. The second one (GLUE) is the sales of gluestick at
2033 US retail locations. The third (PARTS) is the demand for spare parts
of a major European IT firm, previously studied by
\citet{Syntetos:2012ea}. The first two datasets illustrate variability at
the location level for the same item (SKU), whereas the third one
illustrates variability at the item level. All time series are non-negative
integers, some with a large fraction of zeros, and all series of a given
dataset covers the same date range; Table~\ref{tab:summary-statistics}
shows some summary statistics. The ``mean non-zero value'' is the mean
series value, conditional on the value being positive; the ``mean non-zero
inter-period'' is the number of periods between non-zero observations; and
the ``mean sq. coef. of variation'' is the square of the coefficient of
variation ($\mathrm{CV}^2$), which is $\sigma^2/m^2$ for a series with mean
$m$ and standard deviation $\sigma$.

\begin{table}
\caption{Summary statistics of datasets evaluated. The categorization of
  demand patterns into ``smooth'' (high non-zero demand rate, low
  $\mathrm{CV}^2$), ``erratic'' (both high), ``intermittent'' (both low)
  and ``lumpy'' (low demand rate, high $\mathrm{CV}^2$) follows
  \citet{Syntetos:2005:categorization}. The chosen datasets cover a broad
  mix of patterns.}
\label{tab:summary-statistics}
\footnotesize
  \renewcommand{\arraystretch}{0.85}%
  \begin{tabular*}{\columnwidth}{@{\extracolsep{\fill}}lrrr@{}}
    \toprule
                                      & RAID  & GLUE   & PARTS  \\
    \midrule
    Number of time series             & 24    & 2033   & 3055   \\
    Sampling period                   & Weekly& Weekly & Monthly\\
    Nb. of observations per series    & 66    & 79     & 48     \\
    Initial training set duration     & 53    & 65     & 24     \\
    Mean non-zero value               & 6.31  & 1.53   & 23.73  \\
    Mean non-zero inter-period        & 1.15  & 4.02   & 3.82   \\
    Mean sq. coef. of variation $(\mathrm{CV}^2)\!\!\!\!\!\!\!\!$
                                      & 0.39  & 0.29   & 1.29   \\
    \midrule
    \% Smooth                         & 57\%  & 1\%    & 4\%    \\
    \% Erratic                        & 27\%  & 0\%    & 15\%   \\
    \% Intermittent                   & 17\%  & 90\%   & 14\%   \\
    \% Lumpy                          & 0\%   & 9\%    & 67\%   \\
    \bottomrule
  \end{tabular*}
\vspace*{-2mm}
\end{table}

Model performance is evaluated by the out-of-sample forecasting accuracy
over the horizon $h$ (where $h$ varies from 1 to 12 periods), measured
according to the negative log-likelihood (NLL) per period, relative mean
squared error (MSE) and relative mean absolute error (MAE). To reduce the
scale dependence of the MSE and MAE, the MSE is normalized by the in-sample
variance and the MAE is normalized by the in-sample mean absolute deviation
from the in-sample mean. We evaluate performance by a sequential
re-training procedure that alternates between model training and testing,
moving at each iteration the first observation of the (previous) test set
to the end of the (new) training set. This simulates the action of a
decision-maker acting in real-time, retraining models as new information
becomes available. All reported results are averages of out-of-sample
performance under this procedure. The initial training set durations for
each dataset are given in Table~\ref{tab:summary-statistics}.

\subsection{Benchmark Models}

We compare the forecasting performance of the proposed H-NBSS model against
the following benchmarks: (\emph{i})~\citeauthor{Croston:1972}'s
\citeyearpar{Croston:1972} method, (\emph{ii})~simple exponential smoothing
(E-S) with additive errors and an automatically-adjusted smoothing
constant, and (\emph{iii})~the damped dynamic model with negative binomial
observations of \citet{Snyder:2012:forecasting}. The first two approaches
are as implemented by the corresponding functions in the R
\texttt{forecast} package \citep{Hyndman:2013:forecast}. Since they provide
point forecasts only, we evaluate predictive distributions under two
alternatives: a Gaussian distribution, with variance given by the variance
of training residuals, or a Poisson distribution. In both cases, the mean
is given by the point forecast.

\subsection{Performance Results}

\begin{table*}[t!]
  \caption{Forecasting performance at various horizons. For all measures, a
    lower value indicates a higher accuracy; best results are bolded.}  
  \footnotesize
  \renewcommand{\arraystretch}{0.85}%
    \begin{tabular*}{\textwidth}{@{\extracolsep{\fill}}lrrrrrrrrrrrr@{}}
    \toprule
          &       & \multicolumn{3}{c}{Normalized NLL} & \multicolumn{1}{c}{} & \multicolumn{3}{c}{Relative MSE} & \multicolumn{1}{c}{} & \multicolumn{3}{c}{Relative MAE} \\
    \midrule
    \multicolumn{13}{c}{Dataset: RAID} \\
    \midrule
    Forecast Horizon (periods)      &       & \multicolumn{1}{c}{1} & \multicolumn{1}{c}{4} & \multicolumn{1}{c}{8} & \multicolumn{1}{c}{} & \multicolumn{1}{c}{1} & \multicolumn{1}{c}{4} & \multicolumn{1}{c}{8} & \multicolumn{1}{c}{} & \multicolumn{1}{c}{1} & \multicolumn{1}{c}{4} & \multicolumn{1}{c}{8} \\
    \cmidrule(lr){3-5}\cmidrule(lr){7-9}\cmidrule(lr){11-13}
    Croston (Gaussian) &       & 2.408 & 2.461 & 2.523 &       & 0.879 & 1.003 & 1.133 &       & 0.900 & 0.969 & 1.024 \\
    Croston (Poisson) &       & 2.457 & 2.620 & 2.827 &       & 0.879 & 1.003 & 1.133 &       & 0.900 & 0.969 & 1.024 \\
    E-S Additive (Gaussian) &       & 2.284 & 2.356 & 2.492 &       & 0.708 & 0.830 & 1.047 &       & 0.824 & 0.883 & 0.981 \\
    E-S Additive (Poisson) &       & 2.312 & 2.447 & 2.775 &       & 0.708 & 0.830 & 1.047 &       & 0.824 & 0.883 & 0.981 \\
    Snyder-Ord-Beaumont &       & 2.285 & 2.340 & 2.384 &       & 0.749 & 0.822 & 0.869 &       & 0.836 & 0.871 & 0.893 \\
    H-NBSS w/o Seas (Laplace) &       & 2.251 & 2.275 & 2.378 &       & 0.677 & 0.732 & 0.837 &       & 0.811 & 0.836 & 0.876 \\
    H-NBSS w/o Seas (MCMC) &       & 2.286 & 2.380 & 2.595 &       & 0.728 & 0.847 & 1.026 &       & 0.835 & 0.893 & 0.972 \\
    H-NBSS w/  Seas (Laplace) &       & \textbf{2.228} & \textbf{2.219} & \textbf{2.265} &       & 0.656 & 0.669 & 0.714 &       & 0.800 & \textbf{0.808} & \textbf{0.814} \\
    H-NBSS w/  Seas (MCMC) &       & 2.231 & 2.278 & 2.362 &       & \textbf{0.634} & \textbf{0.654} & \textbf{0.693} &       & \textbf{0.796} & 0.821 & 0.849 \\
    \midrule
    \multicolumn{13}{c}{Dataset: GLUE} \\
    \midrule
    Forecast Horizon (periods)      &       & \multicolumn{1}{c}{1} & \multicolumn{1}{c}{4} & \multicolumn{1}{c}{8} & \multicolumn{1}{c}{} & \multicolumn{1}{c}{1} & \multicolumn{1}{c}{4} & \multicolumn{1}{c}{8} & \multicolumn{1}{c}{} & \multicolumn{1}{c}{1} & \multicolumn{1}{c}{4} & \multicolumn{1}{c}{8} \\
    \cmidrule(lr){3-5}\cmidrule(lr){7-9}\cmidrule(lr){11-13}
    Croston (Gaussian) &       & 1.238 & 1.256 & 1.269 &       & 1.126 & 1.141 & 1.144 &       & 1.029 & 1.032 & 1.031 \\
    Croston (Poisson) &       & 0.881 & 0.880 & 0.880 &       & 1.126 & 1.141 & 1.144 &       & 1.029 & 1.032 & 1.031 \\
    E-S Additive (Gaussian) &       & 1.241 & 1.254 & 1.267 &       & 1.123 & 1.132 & 1.130 &       & 0.997 & 0.998 & 0.994 \\
    E-S Additive (Poisson) &       & 0.871 & 0.868 & 0.866 &       & 1.123 & 1.132 & 1.130 &       & 0.997 & 0.998 & 0.994 \\
    Snyder-Ord-Beaumont &       & 0.872 & 0.876 & 0.889 &       & 1.122 & 1.137 & 1.139 &       & 0.982 & 0.985 & 0.985 \\
    H-NBSS w/o Seas (Laplace) &       & \textbf{0.830} & \textbf{0.827} & \textbf{0.825} &       & 1.112 & \textbf{1.124} & 1.127 &       & \textbf{0.975} & \textbf{0.976} & \textbf{0.977} \\
    H-NBSS w/o Seas (MCMC) &       & 0.835 & 0.832 & 0.829 &       & \textbf{1.111} & \textbf{1.124} & \textbf{1.126} &       & 1.002 & 1.006 & 1.006 \\
    \midrule
    \multicolumn{13}{c}{Dataset: PARTS} \\
    \midrule
    Forecast Horizon (periods)      &       & \multicolumn{1}{c}{1} & \multicolumn{1}{c}{4} & \multicolumn{1}{c}{8} & \multicolumn{1}{c}{} & \multicolumn{1}{c}{1} & \multicolumn{1}{c}{4} & \multicolumn{1}{c}{8} & \multicolumn{1}{c}{} & \multicolumn{1}{c}{1} & \multicolumn{1}{c}{4} & \multicolumn{1}{c}{8} \\
    \cmidrule(lr){3-5}\cmidrule(lr){7-9}\cmidrule(lr){11-13}
    Croston (Gaussian) &       & 68.28 & 78.87 & 98.66 &       & \textbf{132.39} & \textbf{150.31} & \textbf{155.10} & \textbf{} & 1.982 & 2.212 & 2.332 \\
    Croston (Poisson) &       & 14.53 & 15.88 & 16.76 &       & \textbf{132.39} & \textbf{150.31} & \textbf{155.10} & \textbf{} & 1.982 & 2.212 & 2.332 \\
    E-S Additive (Gaussian) &       & 69.07 & 78.11 & 81.19 &       & 132.50 & 150.45 & 155.26 &       & 1.879 & 2.116 & 2.238 \\
    E-S Additive (Poisson) &       & 14.17 & 16.52 & 18.49 &       & 132.50 & 150.45 & 155.26 &       & 1.879 & 2.116 & 2.238 \\
    Snyder-Ord-Beaumont &       & 8.80  & 19.48 & 38.24 &       & 132.51 & 150.48 & 155.28 &       & \textbf{1.833} & \textbf{2.063} & \textbf{2.180} \\
    H-NBSS w/o Seas (Laplace) &       & \textbf{4.21} & \textbf{4.51} & \textbf{4.91} & \textbf{} & 132.48 & 150.43 & 155.23 &       & 1.866 & 2.097 & 2.217 \\
    \bottomrule
    \end{tabular*}
    \label{tab:forecasting-results}
\end{table*}%

\begin{figure}
\resizebox{\columnwidth}{!}{\includegraphics{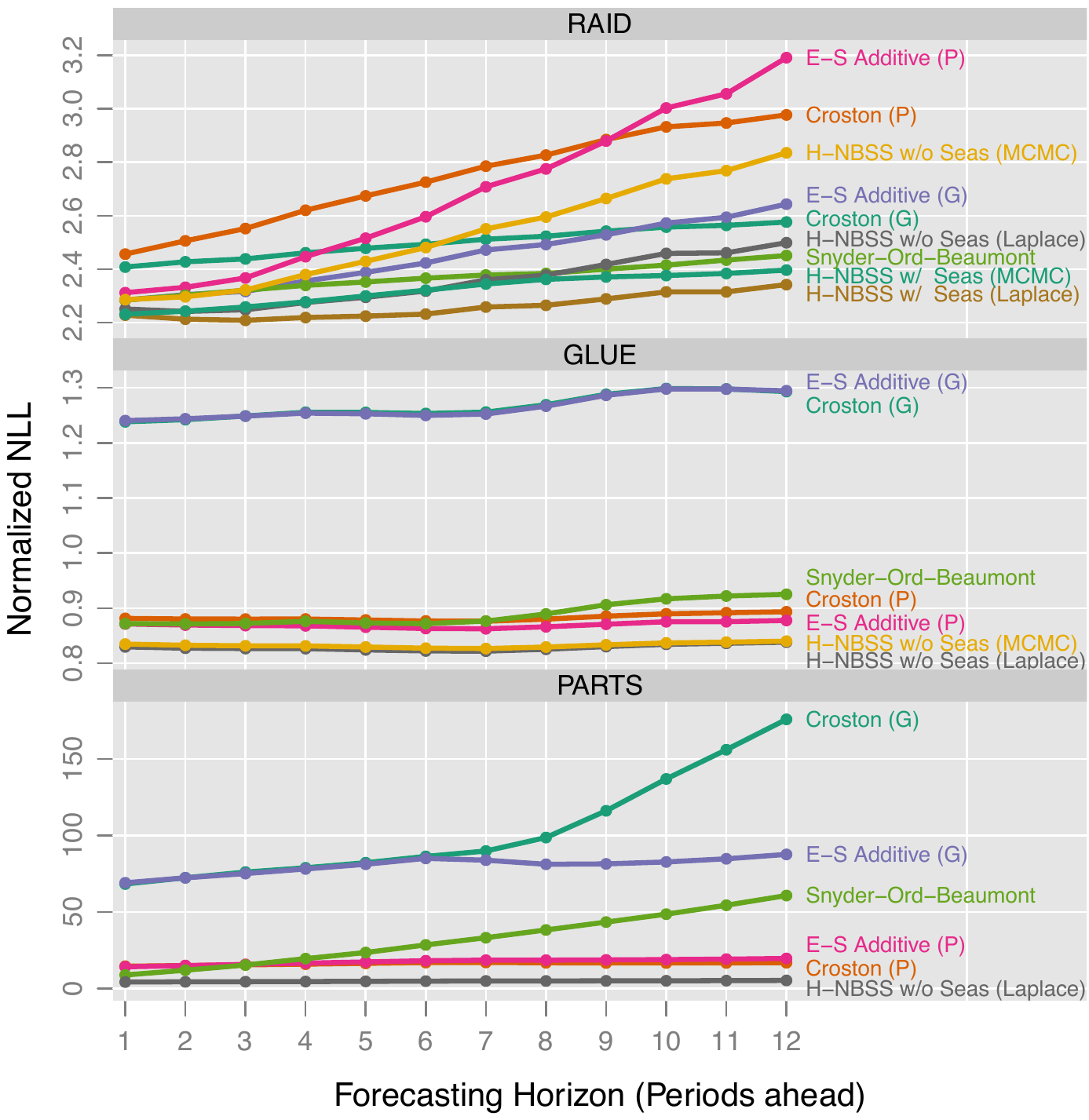}}
\vspace*{-7mm}
\caption{Average out-of-sample NLL for all datasets as a function of the
  forecasting horizon. The proposed H-NBSS model with the Laplace
  approximation exhibits the best performance.}
\label{fig:OOS-NLL}
\end{figure}

\begin{figure*}[t]
\resizebox{\textwidth}{!}{\includegraphics{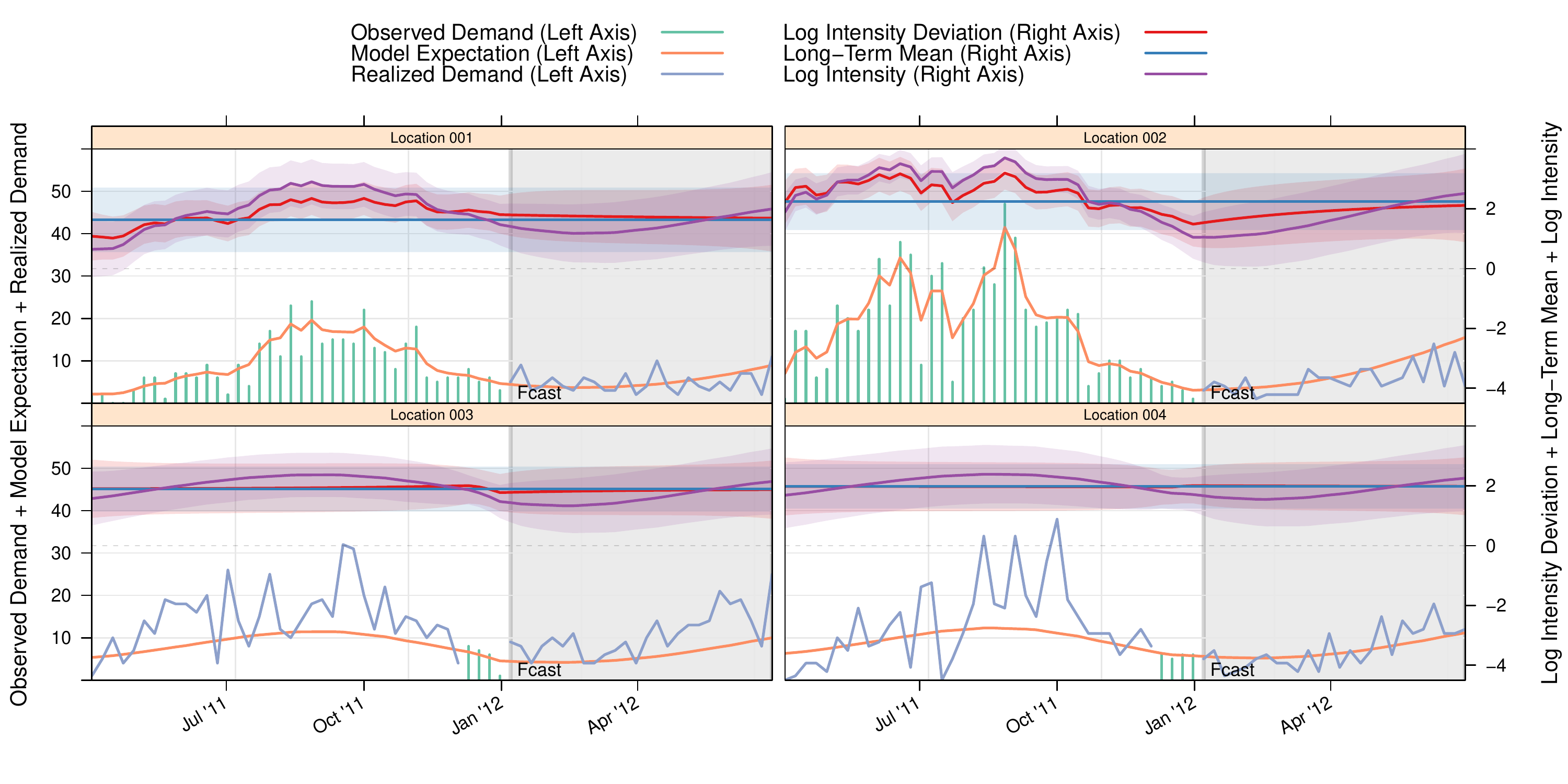}}%
\vspace*{-5.5mm}
\resizebox{\textwidth}{!}{\includegraphics{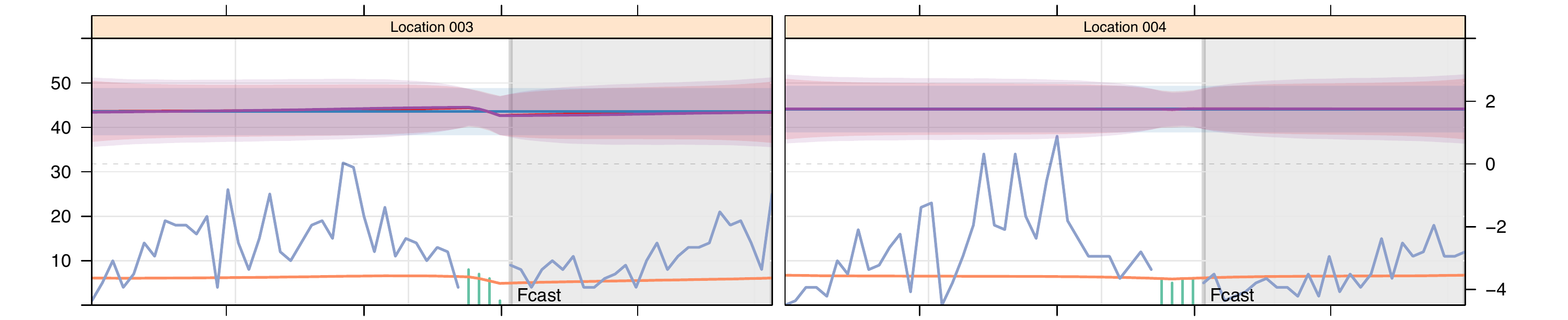}}
\vspace*{-5mm}
\caption{\textbf{Top 4 panels:} Example of information sharing in the
  hierarchical model. The time series for 4 stores of the RAID dataset are
  included in the group, the first two with a full history (green vertical
  lines) and the bottom two with only four observations of history each
  (which appear just before the start of the forecasting period). The
  realized observations are indicated by the solid blue lines and the
  expectation under the model distribution is the orange line; we note that
  the model deduces sensible seasonalities over the forecasting horizon for
  Locations~3 and~4 even with only four observations in their respective
  histories, and can reasonably ``backcast'' over the missing history as
  well. \textbf{Bottom 2 panels:} Results with an independent model for
  each series (no sharing); with only four observations in the history, the
  models cannot do much better than fit a constant.}
\label{fig:information-sharing}
\end{figure*}

Seasonalities are significant on the RAID dataset, which are incorporated
into the H-NBSS through explanatory variables; for this dataset, we report
H-NBSS results both without and with seasonalities, for both the Laplace
and MCMC approximate inference. Seasonalities are omitted from GLUE and
PARTS for space reasons since they yield very similar performance to the
model without seasonal effects. Moreover, H-NBSS results in this section
consider each dataset series independently of the others (i.e. groups of
size~1). The benefits of hierarchy are examined in the next section.
Out-of-sample performance results for all datasets and models at selected
forecasting horizons are given in Table~\ref{tab:forecasting-results}. NLL
results at all horizons appear in Fig.~\ref{fig:OOS-NLL}. We observe that
on the NLL measure (which measures predictive distributional accuracy) the
H-NBSS model consistently yields the best performance, with the Laplace
approximation slightly beating MCMC (both approximations are very
close). The MSE and MAE measures tell a consistent story, the only
exception being the PARTS dataset where Croston very slightly bests the
other approaches in the forecast of the mean (MSE measure). This can be
explained by high proportion of ``lumpy'' series in PARTS (\emph{cf.}
Table~\ref{tab:summary-statistics}), which exhibit high demand variability,
and hence low predictability.

\subsection{Benefits of Hierarchy}

\begin{table}
  \caption{Forecasting performance for the masked series in the RAID
    dataset, for which only four observations are available.}
  \label{tab:hierarchy-benefits}
  \footnotesize
  \renewcommand{\arraystretch}{0.85}%
    \begin{tabular*}{\columnwidth}{@{\extracolsep{\fill}}lccc@{}}
    \toprule
          & Normalized & Relative & Relative \\
          & NLL        & MSE      & MAE      \\
    \midrule
    Independent models & 79.04 &  7.69 &  2.13 \\
    Hierarchical model & \textbf{72.77} & \textbf{4.13} & \textbf{1.75} \\
    \bottomrule
    \end{tabular*}
\end{table}

We close this section by outlining the benefits of the hierarchical
structure in H-NBSS. The major advantage of sharing information across
several series in a group lies in the ability to increase ``statistical
strength'', in particular for series for which little history is
available. We illustrate this in Fig.~\ref{fig:information-sharing}, which
shows that when groups of time series can share information, useful
patterns can be learned even for series with very short histories (here,
four observations). In contrast, with no sharing, the model cannot do much
better than fit a constant.

These results translate quantitatively in
Table~\ref{tab:hierarchy-benefits}. On 20 of the 24 series of the RAID
dataset, we provided only the four observations before the forecasting
horizon; for the other 4 series, we provided the full history. The table
contrasts the performance of a H-NBSS model trained separately for each
series (``independent models''), versus a single H-NBSS model grouping the
24 series together (``hierarchical model'') and reports average performance
only on the 20 masked series. There is a dramatic gain in performance
attributable to sharing in the hierarchical model: seasonalities learned on
the four complete series transfer to the incomplete ones.

%%%%%%%%%%%%%%%%%%%%%%%%%%%%%%%%%%%%%%%%%%%%%%%%%%%%%%%%%%%%%%%%%%%%%%%%%%%%%%%

\section{Conclusion}

This paper introduced a modeling methodology for groups of related time
series of counts, such as the small-integer series frequently encountered
in supply chain operations. We outlined the sizable accuracy gains possible
through jointly modeling several time series in a hierarchical Bayesian
framework and presented an effective approximate inference algorithm to
make the H-NBSS model useful in practice. Future work should investigate
other tradeoffs on the accuracy--computational cost spectrum, such as the
INLA approach \citep{Rue:2009:approximate}, recently revisited by
\citet{Han:2013:non-factorized}. Beyond its increased accuracy, the
\mbox{H-NBSS} model can provide real-world benefits in applications
where count data dominate, for instance by supplying useful forecasts for
new stores with very little (or no) history and improving the efficiency of
inventory management policies with better distributions of future demand.

%%%%%%%%%%%%%%%%%%%%%%%%%%%%%%%%%%%%%%%%%%%%%%%%%%%%%%%%%%%%%%%%%%%%%%%%%%%%%%%

%% % Acknowledgements should only appear in the accepted version. 
\section*{Acknowledgments} 

The author wishes to thank his colleagues at ApSTAT Technologies and JDA
Software for constructive discussions, as well as the anonymous reviewers
for their insight and useful comments.

%% \textbf{Do not} include acknowledgements in the initial version of
%% the paper submitted for blind review.
%% 
%% If a paper is accepted, the final camera-ready version can (and
%% probably should) include acknowledgements. In this case, please
%% place such acknowledgements in an unnumbered section at the
%% end of the paper. Typically, this will include thanks to reviewers
%% who gave useful comments, to colleagues who contributed to the ideas, 
%% and to funding agencies and corporate sponsors that provided financial 
%% support.  

\bibliography{slow_movers_bib}
\bibliographystyle{icml2014}

\pagebreak

\appendix
\section{Process Hyper-Priors}
\label{app:process-hyperpriors}

The following hyperpriors are used for the hierarchical model described in
section~2.3:
\begin{align*}
    \bar{\alpha}    &\sim U(0.001,0.1),                                 &
    \mu_{\mu}       &\sim \Normal(0, 2^2),                              \\
    \tau_{\mu}      &\sim U(1, 10),                                     & 
    \kappa_\tau     &\sim U(5, 10),                                     \\
    \beta_\tau      &\sim U(2, 25),                                     & 
    \kappa_{0,\tau} &\sim U(1, 5),                                      \\
    \beta_{0,\tau}  &\sim U(1, 10),                                     & 
    \kappa_\theta   &\sim U(5, 10),                                     \\
    \beta_\theta    &\sim U(2, 25),                                     & 
    \phi_{+}        &\sim U(1, 600),                                    \\
    \phi_{-}        &\sim U(1, 50),                                     &
    \bar{\btheta}   &\sim \Normal(0, 1).
\end{align*}

\section{Precision Matrix for Hierarchical GMRF Prior}

The hierarchical model described in section 2.3 gives rise to a conditional
Gaussian Markov random field (GMRF) prior over the global level of
mean-reversion $\mu_\mu$, the series-specific levels of mean-reversion
$\mu_{\ell}, \ell=1,\ldots,L$, and the latent process log-means $\{
\eta_{\ell,t} \}$. The GMRF prior structure for two time series is
illustrated in the following graph:

\begin{tikzpicture}
  \node[latent](mumu){$\mu_\mu$};
  \node[latent, above=2.0 of mumu, xshift=+2cm](mu1){$\mu_1$};
  \node[latent, below=0.8 of mumu, xshift=+2cm](mu2){$\mu_2$};

  \node[latent, below=1.0 of mu1  ](eta11){$\eta_{1,1}$};
  \node[latent, right=0.5 of eta11](eta12){$\eta_{1,2}$};
  \node[latent, right=0.5 of eta12](eta13){$\eta_{1,3}$};
  \node[const , right=0.5 of eta13](eta1D){$\;\cdots\;$};
  \node[latent, right=0.5 of eta1D](eta1T){$\eta_{1,T}$};

  \node[latent, below=1.0 of mu2  ](eta21){$\eta_{2,1}$};
  \node[latent, right=0.5 of eta21](eta22){$\eta_{2,2}$};
  \node[latent, right=0.5 of eta22](eta23){$\eta_{2,3}$};
  \node[const , right=0.5 of eta23](eta2D){$\;\cdots\;$};
  \node[latent, right=0.5 of eta2D](eta2T){$\eta_{2,T}$};

  \edge[-]{mumu}{mu1,mu2};

  \edge[-]{mu1}{eta11,eta12,eta13,eta1D,eta1T};
  \edge[-]{eta11}{eta12};
  \edge[-]{eta12}{eta13};
  \edge[-]{eta13}{eta1D};
  \edge[-]{eta1D}{eta1T};

  \edge[-]{mu2}{eta21,eta22,eta23,eta2D,eta2T};
  \edge[-]{eta21}{eta22};
  \edge[-]{eta22}{eta23};
  \edge[-]{eta23}{eta2D};
  \edge[-]{eta2D}{eta2T};
\end{tikzpicture}

In a GMRF, an edge in the graphical model corresponds to a non-zero entry
in the precision matrix of the joint distribution over all
variables. Hence, the precision matrix is very sparse: it has block
diagonal structure, where each block corresponds to a single series. In the
two-series example, assuming that each series has 4 observations, we have
the following precision matrix:

\setcounter{MaxMatrixCols}{20}
\renewcommand{\arraycolsep}{1.5pt}
\renewcommand{\arraystretch}{1.25}

\begin{widetext}
\begin{equation*}
Q =
\footnotesize
\begin{bmatrix}[ccccc:ccccc:c]
 \tau_1 & -\tau_1 \phi_1 & 0 & 0 & \tau_1 \tilde{\phi}_1 & 0 & 0 & 0 & 0 & 0 & 0 \\
 -\tau_1 \phi_1 & \tau_1 \left(\phi_1^2+1\right) & -\tau_1 \phi_1 & 0 & -\tau_1 \tilde{\phi}_1^2 & 0 & 0 & 0 & 0 & 0 & 0 \\
 0 & -\tau_1 \phi_1 & \tau_1 \left(\phi_1^2+1\right) & -\tau_1 \phi_1 & -\tau_1 \tilde{\phi}_1^2 & 0 & 0 & 0 & 0 & 0 & 0 \\
 0 & 0 & -\tau_1 \phi_1 & \tau_1 & \tau_1 \tilde{\phi}_1 & 0 & 0 & 0 & 0 & 0 & 0 \\
 \tau_1 \tilde{\phi}_1 & -\tau_1 \tilde{\phi}_1^2 & -\tau_1 \tilde{\phi}_1^2 & \tau_1 \tilde{\phi}_1 & \tau_{\mu_1}+ \tau_1 \psi_{1,T} & 0 & 0 & 0 & 0 & 0 & -\tau_{\mu_1} \\ \hdashline
 0 & 0 & 0 & 0 & 0 & \tau_2 & -\tau_2 \phi_2 & 0 & 0 & \tau_2 \tilde{\phi}_2 & 0 \\
 0 & 0 & 0 & 0 & 0 & -\tau_2 \phi_2 & \tau_2 \left(\phi_2^2+1\right) & -\tau_2 \phi_2 & 0 & -\tau_2 \tilde{\phi}_2^2 & 0 \\
 0 & 0 & 0 & 0 & 0 & 0 & -\tau_2 \phi_2 & \tau_2 \left(\phi_2^2+1\right) & -\tau_2 \phi_2 & -\tau_2 \tilde{\phi}_2^2 & 0 \\
 0 & 0 & 0 & 0 & 0 & 0 & 0 & -\tau_2 \phi_2 & \tau_2 & \tau_2 \tilde{\phi}_2 & 0 \\
 0 & 0 & 0 & 0 & 0 & \tau_2 \tilde{\phi}_2 & -\tau_2 \tilde{\phi}_2^2 & -\tau_2 \tilde{\phi}_2^2 & \tau_2 \tilde{\phi}_2 & \tau_{\mu_2}+ \tau_2 \psi_{2,T} & -\tau_{\mu_2} \\ \hdashline
 0 & 0 & 0 & 0 & -\tau_{\mu_1} & 0 & 0 & 0 & 0 & -\tau_{\mu_2} & \tau_{\mu_1}+\tau_{\mu_2}+\tau_{\mu_\mu}  \\
\end{bmatrix}
\end{equation*}
\end{widetext}
where $T=4$ (the number of periods), $\tilde{\phi}_1 = \phi_1-1$,
$\tilde{\phi}_2 = \phi_2-1$, $\psi_{1,T} \equiv T - 2(T-1)\phi_1 +
(T-2)\phi_1^2$, $\psi_{2,T} \equiv T - 2(T-1)\phi_2 + (T-2)\phi_2^2$.  The
block structure is emphasized with dashed lines.  The determinant of this
matrix is $\tau_{\mu_\mu} \bigl(\tau_{\mu_1} \tau_1^T (\phi_1^2-1)\bigr)
\bigl(\tau_{\mu_2} \tau_2^T (\phi_2^2-1)\bigr)$, which is useful for computing the
probability of a variable configuration.

\end{document}